\documentclass[10pt,twocolumn,letterpaper]{article}
\usepackage[final]{cvpr}

\usepackage[numbers,sort&compress]{natbib}
\usepackage{amsmath,amssymb,amsfonts}
\usepackage{algorithmic}
\usepackage{graphicx}
\usepackage{textcomp}
\usepackage{multirow}
\usepackage{mathtools}
\usepackage{booktabs}
\usepackage{diagbox}
\usepackage{xspace}
\usepackage{bbm}
\usepackage[table,x11names,dvipsnames]{xcolor}
\usepackage{subcaption}
\usepackage{etoolbox}
\usepackage[most]{tcolorbox}

\definecolor{LightGrey}{rgb}{0.9,0.9,0.9}
\definecolor{pastelyellow}{HTML}{d1af42}
\definecolor{pastelblue}{HTML}{a1b5cf}

\newcommand\tstrut{\rule{0pt}{2.2ex}}

\newcommand\thindots{\makebox[1em][c]{.\hfil.\hfil.}}

\newcommand\AppendixA{%
  \xdef\presupfigures{\arabic{figure}}
  \xdef\presupsections{\arabic{section}}
  \renewcommand\thefigure{A-\fpeval{\arabic{figure}-\presupfigures}}
  \renewcommand\thesection{A-\fpeval{\arabic{section}-\presupsections}}
  \renewcommand{\thetable}{A-\Roman{table}}
  \renewcommand{\theequation}{A-\arabic{equation}}
}

\definecolor{grdpurp}{HTML}{B5739D}

\definecolor{cvprblue}{rgb}{0.21,0.49,0.74}
\usepackage[pagebackref,breaklinks,colorlinks,allcolors=cvprblue]{hyperref}

\def\paperID{2} 
\def\confName{CVPR}
\def\confYear{2026}

\title{MIMIC: Multimodal Inversion for Model Interpretation and Conceptualization}

\author{Animesh Jain, Alexandros Stergiou \\
University of Twente, The Netherlands \\
{\tt\small a.jain-2@alumnus.utwente.nl, a.g.stergiou@utwente.nl}
}

\begin{document}
\maketitle

\begin{abstract}
Vision Language Models (VLMs) encode multimodal inputs over large, complex, and difficult-to-interpret architectures, which limit transparency and trust. We propose a Multimodal Inversion for Model Interpretation and Conceptualization (MIMIC) framework that inverts the internal encodings of VLMs. MIMIC uses a joint VLM-based inversion and a feature alignment objective to account for VLM's autoregressive processing. It additionally includes a triplet of regularizers for spatial alignment, natural image smoothness, and semantic realism. We evaluate MIMIC both quantitatively and qualitatively by inverting visual concepts across a range of free-form VLM outputs of varying length. Reported results include both standard visual quality metrics and semantic text-based metrics. To the best of our knowledge, this is the first model inversion approach addressing visual interpretations of VLM concepts. Project page: \href{https://anaekin.github.io/MIMIC}{https://anaekin.github.io/MIMIC}
\end{abstract}

\section{Introduction}

\noindent
Vision-Language Models (VLMs) have demonstrated impressive capabilities in numerous tasks. Despite their ability to encode multiple modalities, we still face difficulties in determining whether models' decisions are grounded on internal post-training reasoning~\cite{chuang2024faithlm, chen2024selfie, ghandeharioun2024patchscopes} or are instead interpolated from memorized training examples~\cite{grosse2023influencefunctions, meng2023factualassociations}.

Uncovering visual explanations for model decisions has been the focus of many research works. Methods have visualized image region attributions~\cite{gradcam,sundararajan2017axiomatic}, saliency and hidden activations~\cite{raghu2021vision}, traced information flow, or editing representations in LLMs~\cite{meng2023factualassociations,zhao2025gradeclip,golovanevsky2025vlmsnotice,ghandeharioun2024patchscopes}. These methods, however, are primarily unimodal and rely on gradient access, auxiliary decoders, or architecture-specific modifications.

\begin{figure}[t]
    \centering
    \includegraphics[width=.98\linewidth]{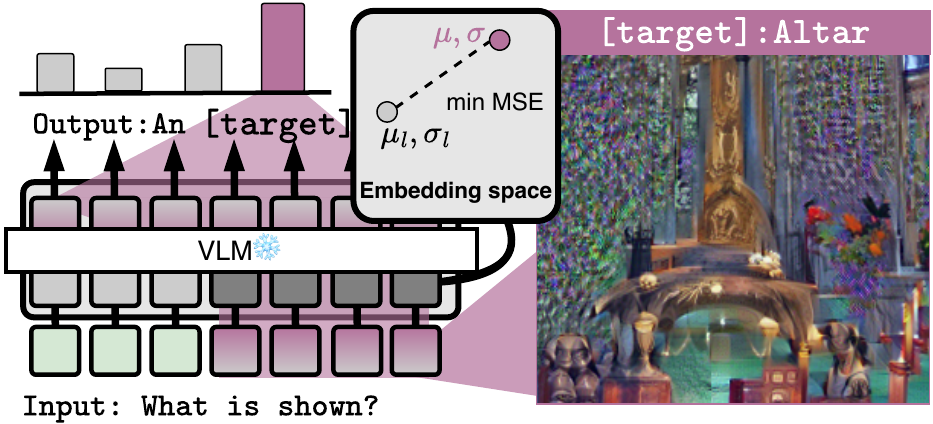}
    \vspace*{-0.4em}
    \caption{\textbf{Multimodal Inversion for Model Interpretation and Conceptualization (MIMIC)} inverts VLMs by synthesizing visual inputs that best correspond to VLM tokens and internal embeddings. The synthesized images represent the dominant visual features associated with predicted tokens.}
    \label{fig:teaser}
    \vspace{-1.5em}
\end{figure}

In this paper, we address this interpretability gap by optimizing visual inputs for VLM tokens with a Multimodal Inversion for Model Interpretation and Conceptualization (MIMIC) framework, shown in~\cref{fig:teaser}. MIMIC extends current unimodal inversion to autoregressive multimodal models. We use VLM logits as our optimization targets, along with a guidance objective to align the distributions of visual token encodings. Regularizers are added to promote smoothness, total variance, and alignment across distributions. We validate MIMIC's effectiveness using a diverse set of evaluation metrics for semantic alignment with textual prompts, perceptual quality, and embedding similarity.

Our contributions are: i) A model inversion objective that can optimize visual inputs from VLM logits. ii) MIMIC, a general visual interpretability approach, which, to the best of our knowledge, is the first attempt at inverting learned VLM visual features corresponding to tokens. iii) We show that MIMIC can invert VLM text semantics of different lengths to high-fidelity images.

\begin{figure*}[ht]
    \centering
    \includegraphics[width=\linewidth,trim={3cm 0 0 0},clip]{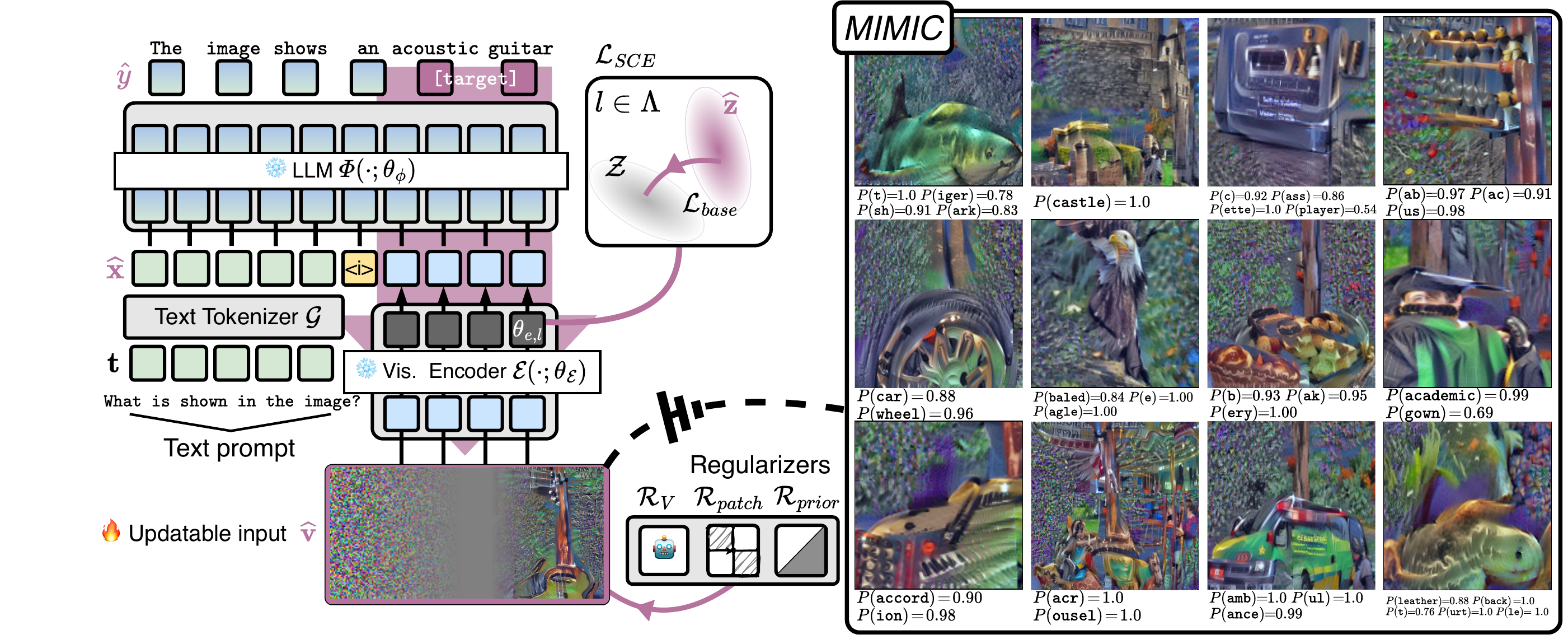}
    \caption{\textbf{MIMIC inversion} iteratively optimizes an updatable input $\color{grdpurp}\widehat{\mathbf{v}}$ with an adapted cross-entropy loss, $\mathcal{L}_{SCE}$, to maximize the probability distribution of \texttt{[target]} VLM token(s), and a base feature loss, $\mathcal{L}_{base}$, to match layer statistics to target mean and variance within the distribution manifold for vision tokens. Regularizers $\mathcal{R}$ are added to promote variance consistency, visual coherence across tokens, and perceptual quality. Optimized visual inputs are shown from inverting visual-instruct-tuned LLaMA3-8B alongside the per-\texttt{[target]}-token probability $P(\texttt{[target]})$.}
    \label{fig:main}
    \vspace*{-1.2em}
\end{figure*}

\section{Related Work}
\label{chap:related_work}

\noindent
\textbf{Feature attribution} methods localize feature relevance. Class Activation Maps (CAM)-based methods~\cite{zhou2015learning,gradcam,wang2020scorecam} propagate local activations of class-relevant regions in vision models. Pixel-level attributions, such as LRP~\cite{lrp}, DeepLIFT~\cite{shrikumar2019learning}, and IG~\cite{sundararajan2017axiomatic}, visualize gradient-based signals given target outputs. Although attribution methods are widely used to explain CNN/ViT decisions on unimodal supervised tasks, they cannot be directly applied to autoregressive multimodal models. In contrast to these feature-localization approaches, we offer a general inversion method to discover visual encodings corresponding to (semantic information of) VLM tokens.

\noindent
\textbf{Model inversion} methods~\cite{yin2020deepinversion,hatamizadeh2022gradvit,ghiasi2022plugin,stergiou2021mind} synthesize inputs to visually interpret encodings of models. Early approaches relied on Gradient Ascent (GA) by maximizing activations of specific neurons~\cite {yosinski2015understanding}. Later methods relied on generative priors~\cite{nguyen2017plug}. More recently,~\cite{yin2020deepinversion} synthesized realistic images with CNN BN statistics, while~\cite{hatamizadeh2022gradvit,ghiasi2022plugin,stergiou2023leaping} adapted gradient-based inversion to ViTs~\cite{dosovitskiy2021imageworth16x16words}.~\cite{fel2023craft} defined hierarchies of visual concepts, and~\cite{thasarathan2025universal} discovered shared features across models and datasets. Although these inversion techniques have improved the interpretability of vision-only models, their applicability to VLMs remains limited, a problem this paper addresses.

\noindent
\textbf{V/LLM interpretability} methods primarily aim to explain how model encodings relate to abstract concepts.~\cite{chen2024selfie} created text explanations for hidden LLM layers' states, later extended~\cite{ghandeharioun2024patchscopes} by allowing direct modulations over hidden states. Other approaches~\cite{meng2023factualassociations,chuang2024faithlm} focused on modulations in the model’s internal structure with adapters. Works also explored attention and gradient signals. Grad-ECLIP~\cite{zhao2025gradeclip} visualized image and text tokens contributions based on CLIP similarity scores. NOTICE~\cite{golovanevsky2025vlmsnotice} studied image-grounding and object-level reasoning properties within cross-attention heads. Second-Order Lens~\cite{gandelsman2025secondorder} ablated zero-shot accuracy of late layers in CLIP.~\cite{tao2024probing} demonstrated the influence of intermediate layers towards semantic features, while~\cite{oikarinen2024linear,dorszewski2025colors} focused on the alignment of linear layers with concept-relevant features. Despite these efforts, model explanations for VLMs are sparse. We thus take a step forward by inverting VLMs to visualize learned semantic features.

\begin{table*}[t]
\centering
\resizebox{\linewidth}{!}{
\setlength\tabcolsep{1.0pt}
\begin{tabular}{l cccc c cccc c cccc c cccc}
\toprule
\multirow{2.5}{*}{Optimization Objectives} & \multicolumn{4}{c}{LLaMA3-8B} & $\;\;$ & \multicolumn{4}{c}{Mistral-7B} & $\;\;$ & \multicolumn{4}{c}{Vicuna-7B}& $\;\;$ & \multicolumn{4}{c}{Vicuna-13B} \\ \cline{2-5} \cline{7-10} \cline{12-15} \cline{17-20}
& FID $\downarrow$ & LPIPS $\downarrow$ & IS $\uparrow$ & CScr $\uparrow$ && FID $\downarrow$ & LPIPS $\downarrow$ & IS $\uparrow$ & CScr $\uparrow$ && FID $\downarrow$ & LPIPS $\downarrow$ & IS $\uparrow$ & CScr $\uparrow$ && FID $\downarrow$ & LPIPS $\downarrow$ & IS $\uparrow$ & CScr $\uparrow$ \tstrut \\  
\midrule
\rowcolor{LightGrey} \multicolumn{20}{l}{\textit{Partial objectives}} \tstrut \\
Base ($\mathcal{L}_{SCE}$) & 421.63 & 1.71 & 1.21 & 21.84 && 469.42 & 1.88 & 1.03 & 20.64  && 404.51 & 1.68 & 1.10 & 21.44 && 389.45 & 2.14 & 0.95 & 22.38 \\
Base+feat ($\mathcal{L}_{SCE}\!\!+\!\!\mathcal{L}_{base}$) & 317.45 & 0.98 & 3.32 & 26.58  && 340.08 & 1.13 & 2.97 & 26.72 && 326.78 & 1.07 & 3.49 & 27.64 && 277.63 & 1.42 & 3.60 & 25.86 \\
\midrule
\textbf{MIMIC} 
\tstrut & \textbf{178.42} & \textbf{0.73} & \textbf{5.38} & \textbf{29.43}  && \textbf{184.56} & \textbf{0.75} & \textbf{5.18} & \textbf{30.51} && \textbf{162.92} & \textbf{0.83} & \textbf{5.21} & \textbf{29.17} && \textbf{145.19} & \textbf{0.64} & \textbf{5.77} & \textbf{32.50} \tstrut \\
\bottomrule
\end{tabular}
}
\caption{\textbf{Image and semantic quality} comparisons. FID and LPIPS are computed to real images from~\cite{schuhmann2022laion} with results averaged over 1K \texttt{[targets]}. Each loss improves the results. The aggregated objective yields the best overall performance, while each component improves across semantic, perceptual, and distributional metrics. Best results are in \textbf{bold}.}
\label{tab:main1}
\vspace{-1.5em}
\end{table*}

\section{Method}
\label{sec:method}

In this section, we provide a detailed formulation for MIMIC. A conceptual overview of MIMIC is shown in Fig.~\ref{fig:main}. We initialize an \textcolor{grdpurp}{updatable input} $\color{grdpurp}\widehat{\mathbf{v}}$ \( \in \mathbb{R}^{C \times H \times W} \) with \(C\) channels, \(H\) height, and \(W\) width. As VLMs can use multimodal context, we include a text prompt template \( \mathbf{t} \) alongside our updatable input. This takes the form of: \texttt{What is shown in the image?: a.[target]} concept, \texttt{or b.[negative]} concept. For target token \texttt{[tiger]}, this can be \texttt{a.tiger} and \texttt{b.dog}. Text is tokenized by $\mathcal{G}(\mathbf{t})$ to a
sequence of embeddings. Similarly, $\color{grdpurp} \widehat{\mathbf{v}} $ is encoded by $\mathcal{E}$ with $\theta_\mathcal{E}$ parammeters, to embeddings tensor $\mathcal{E}(\color{grdpurp}\widehat{\mathbf{v}}\color{black};\theta_\mathcal{E}) \in \mathtt{R}^{D \times \Omega}$ of \( D \) vision tokens of \(\Omega\) channels. The concatenated text-image context used by the LLM is $\color{grdpurp} \widehat{\mathbf{x}}\color{black} = [ \mathcal{G}(\mathbf{t}),\mathcal{E}(\color{grdpurp}\widehat{\mathbf{v}}\color{black};\theta_\mathcal{E})]$.

\noindent
\textbf{VLM Inversion}. The backbone LLM \( \Phi(\cdot;{\theta_\phi})\), with $\theta_\phi$ frozen params, infers \( \color{grdpurp} \widehat{\mathbf{x}} \) and returns a probabilistic distribution of token logits. Each logit $\color{grdpurp}\widehat{\mathbf{y}}_i\color{black}  = \Phi(\color{grdpurp}\widehat{\mathbf{x}},\widehat{\mathbf{y}}_{<i}\color{black};\theta_\phi)$, where $\color{grdpurp}\widehat{\mathbf{y}}_{<i}\color{black}$ are the previously-generated $i-1$ logits. Due to LLM’s autoregressive nature, generated logits sequences vary in length $|\color{grdpurp}\widehat{\mathbf{y}}\color{black}|$. We define an adapted CE loss $\mathcal{L}_{SCE}$ given the token index $i$ for \texttt{[target]}:
\begin{equation}
    \underset{SCE}{\mathcal{L}}(\color{grdpurp}\widehat{\mathbf{y}}\color{black}) \! = \! - \!\sum\mathbbm{1}(\text{sg}(\widehat{\mathbf{y}}),i,\texttt{[target]}) \text{log}(\color{grdpurp}\widehat{\mathbf{y}}_i\color{black}), 
    \label{eq:main}
\end{equation}
\noindent
where $\mathbbm{1}(\text{sg}(\widehat{\mathbf{y}}),i,\texttt{[target]})$ is the indicator function given sequence index $i$ and target token dictionary index $\texttt{[target]}$, stop gradient $\text{sg}(\widehat{\mathbf{y}})$, and sequence length $|\color{grdpurp}\widehat{\mathbf{y}}\color{black}|$. We minimize this loss to enforce semantic alignment between the updatable image and the target token logits.

\noindent
\textbf{Base Feature Loss}. Although the CE loss provides a strong signal for input updates, it does not guarantee that layer encodings will be within the exact manifold of target tokens' visual features. To align layer encodings $\color{grdpurp}\widehat{\mathbf{z}}\color{black}_l = \mathcal{E}(\color{grdpurp}\widehat{\mathbf{v}}\color{black},\theta_{e,<l})$ with the internal representations of the vision encoder, we approximate the manifold's mean $\mu(\mathcal{Z}_l)$ and variance $\sigma(\mathcal{Z}_l)$ from \texttt{[target]} images given $\theta_{\mathcal{E},l}$ weights, across $l \in \Lambda = \{1,\thindots,L\}$ layers. Internal statistics $\mu(\mathcal{Z}_l)$,$\sigma(\mathcal{Z}_l)$ can also be approximated without sourced images by using a generator, as we demonstrate in \S\ref{appendix:base_feat}. The distribution-matching feature loss takes the form: 
\begin{align}
\begin{split}
     \underset{base}{\mathcal{L}} \!\! = \!\! \sum_{l \in \Lambda} \!\! 
     \Big(\!\|\mu(\color{grdpurp}\widehat{\mathbf{z}}\color{black},\underset{\mathcal{E},l}{\theta}) \!\! - \!\!\mu(\mathcal{Z}_l)\|_2^2  
     \!+\!\|\sigma(\color{grdpurp}\widehat{\mathbf{z}}\color{black},\underset{\mathcal{E},l}{\theta}) \!\! - \!\! \sigma(\mathcal{Z}_l)\|_2^2\Big)
     \label{eq:vit_l2}
\end{split}
\end{align}
\noindent
where \(\Lambda\) denotes ViT layers and $\mathcal{Z}_l$ are the approximated distribution for the \texttt{[target]} feature manifold at layer $l$.

\noindent
\textbf{Regularizers}. VLMs encode multimodal inputs across large, complex feature spaces. We thus improve the optimization objective by including standard regularizers from seminal model inversion methods~\cite{yin2020deepinversion,hatamizadeh2022gradvit,ghiasi2022plugin}. We use \(\mathcal{R}_{patch}\) to smooth color signal variance across ViT tokens and improve the overall uniformity across image patches. We include \(\mathcal{R}_{\text{prior}}\) as a combination of image priors for the total L1/2 feature variance and spatial smoothness \( \mathcal{R}_{TV_1} \) \(\mathcal{R}_{TV_2}\) with a \(\ell_2\)-norm penalty $ \mathcal{R}_{l_2} $ to regularize the range:
\begin{equation}
    \mathcal{R}_{prior}(\color{grdpurp}\widehat{\mathbf{v}}\color{black}) =  \alpha_{1}\mathcal{R}_{TV1}(\color{grdpurp}\widehat{\mathbf{v}}\color{black}) 
    \!+\! \alpha_{2}\mathcal{R}_{TV_2}(\color{grdpurp}\widehat{\mathbf{v}}\color{black}) 
    \!+\! \alpha_{3}\mathcal{R}_{\ell_2}(\color{grdpurp}\widehat{\mathbf{v}}\color{black})
    \label{eq:prior_regulariser}
\end{equation}
where, \(\alpha_{1},\; \alpha_{2}, \;\alpha_{3}\) are the scaling factors used.
\noindent
Regularizer $\mathcal{R}_V$ constraints high-frequency noise (L2) inside each patch to improve realism. The aggregated regularizer \(\mathcal{R}( \color{grdpurp}\mathbf{\widehat{v}}\color{black} ) \) uses \(\beta_{1},\; \beta_{2}\) scaling factors and is defined as:
\begin{equation}
        \mathcal{R}(\color{grdpurp}\mathbf{\widehat{v}}\color{black}) = \beta_{1} \;\mathcal{R}_{V}(\color{grdpurp}\mathbf{\widehat{v}}\color{black}) + \beta_{2} \mathcal{R}_{patch}(\color{grdpurp}\mathbf{\widehat{v}}\color{black}) +\mathcal{R}_{prior}(\color{grdpurp}\mathbf{\widehat{v}}\color{black})
        \label{eq:reg}
\end{equation}
\noindent
\textbf{MIMIC updates} Our objective updates \(\color{grdpurp}\mathbf{\widehat{v}}\color{black}\) per $s$ iterations:
\begin{equation}
    \!\!\!\color{grdpurp}\mathbf{\widehat{v}}\color{black}^{s+1} \!=\! \underset{\color{grdpurp}\mathbf{\widehat{v}^s}\color{black}}{\min}\!{\gamma_1}\!\!\underset{SCE}{\mathcal{L}}
     (\Phi([ \mathcal{G}(\mathbf{t}),\mathcal{E}(\color{grdpurp}\widehat{\mathbf{v}}\color{black};\theta_e)];\theta_\phi) 
    \!\! + \!\! {\gamma_2}\!\underset{base}{\mathcal{L}}
    \!\!\!\! + \!\! \mathcal{R}(\color{grdpurp}\widehat{\mathbf{v}}\color{black})
    \label{eq:final}
\end{equation}
\noindent
it combines (\ref{eq:main}) and (\ref{eq:vit_l2}) losses and the regularizers from (\ref{eq:reg}) with  \(\gamma_{1},\; \gamma_{2}\) factors. The final MIMIC objective inverts VLM encodings relevant to \texttt{[target]} tokens. We note that the method is invariant to the length of \texttt{[target]} and can be used with varying prompted texts $\mathbf{t}$.

\begin{figure}[!ht]
    \centering
    \includegraphics[width=\linewidth]{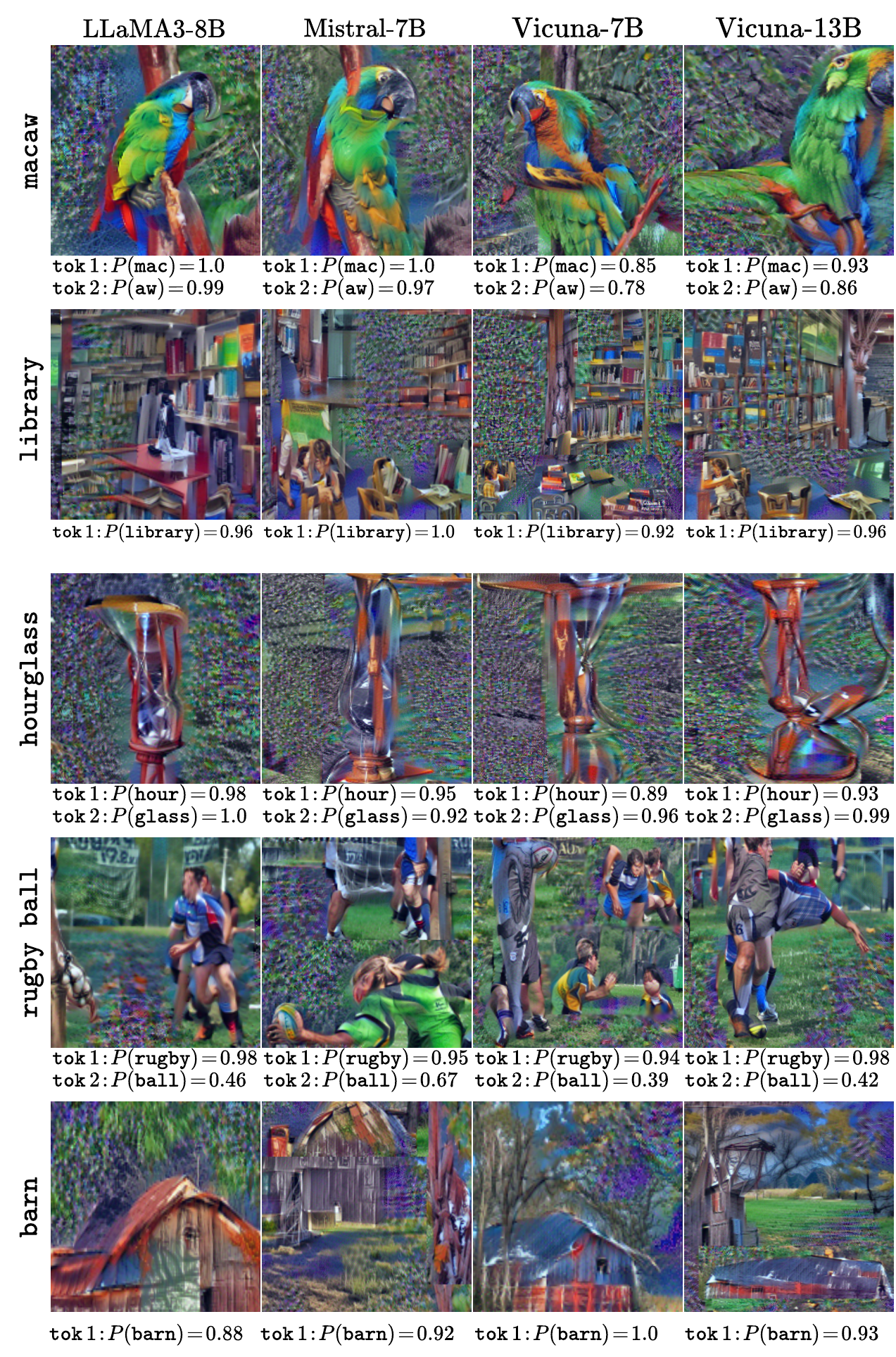}
    \vspace*{-1.2em}
    \caption{\textbf{Qualitative examples of synthesized visual inputs with MIMIC}. Each row optimizes different token logits. The number of tokens corresponding to a \texttt{[target]} semantic concept differs per row.}
    \label{fig:qualitative}
    \vspace{-1em}
\end{figure}

\begin{figure*}[t]
    \centering
    \includegraphics[width=\linewidth]{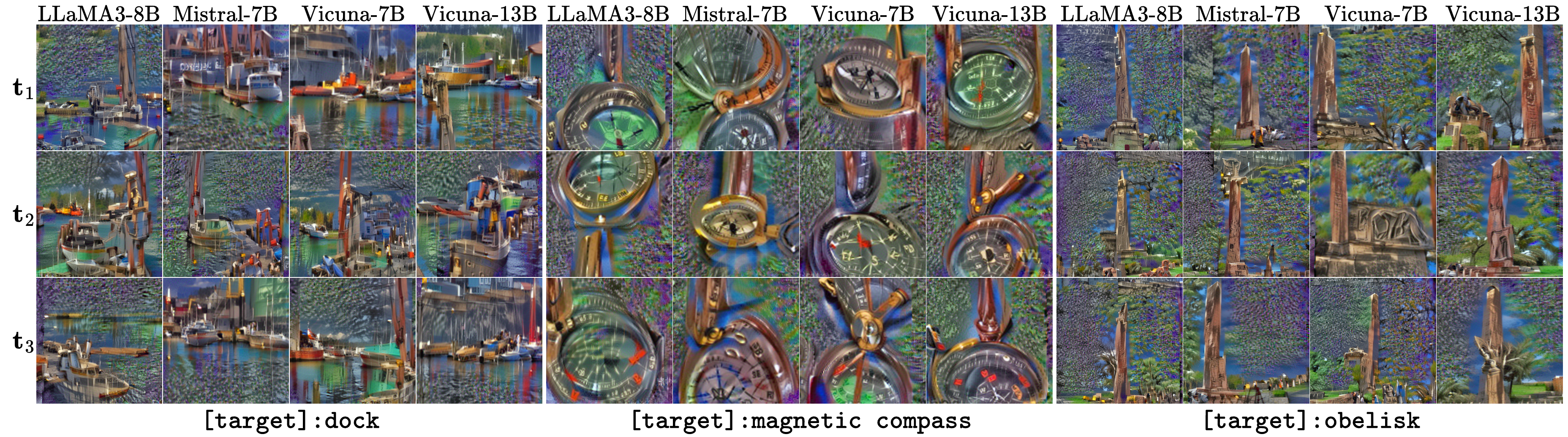}
    \caption{\textbf{Synthesized images over varying text prompts} for \texttt{dock}, \texttt{magnetic compass}, and \texttt{obelisk}. $\mathbf{t}_1$; \texttt{What is shown in the image?a.[target] or b.[negative]}, $\mathbf{t}_2$; \texttt{Does the image show an instance of [target] or [negative]?}, and $\mathbf{t}_3$; \texttt{The image depicts a scene that corresponds to [target] or [negative]?}}
    \label{fig:context-ablations}
    \vspace{-1.2em}
\end{figure*}

\begin{figure*}[t]
\begin{minipage}[b]{.28\linewidth}
    \centering
\resizebox{\linewidth}{!}{
\setlength\tabcolsep{1.0pt}
\begin{tabular}{l  c  c  c  c}
\toprule
Length & BLEU$\uparrow$ & METEOR$\uparrow$ & ROUGE-L$\uparrow$ \\
\midrule
$|\color{grdpurp}\widehat{\mathbf{y}}\color{black}|\!\leq\!2$ & 0.933 & 0.459 & 0.875 \\
$|\color{grdpurp}\widehat{\mathbf{y}}\color{black}|\!\leq\!4$ & 0.928 & 0.465 & 0.880 \\
$|\color{grdpurp}\widehat{\mathbf{y}}\color{black}|\!\leq\!6$ & 0.945 & 0.471 & 0.891 \\
$|\color{grdpurp}\widehat{\mathbf{y}}\color{black}|\!\geq\!7$ & 0.936 & 0.464 & 0.886 \\
\bottomrule
\end{tabular}
}
\captionof{table}{\textbf{Predicted to} \texttt{[target]} \textbf{VLM outputs} across text similarity metrics (BLEU, METEOR, ROUGE-L) on LLaMA3-8B. Results are grouped by text length.}
\label{tab:main2}
\end{minipage}$\;$
\begin{minipage}[b]{.45\textwidth}
    \centering
    \includegraphics[width=\linewidth]{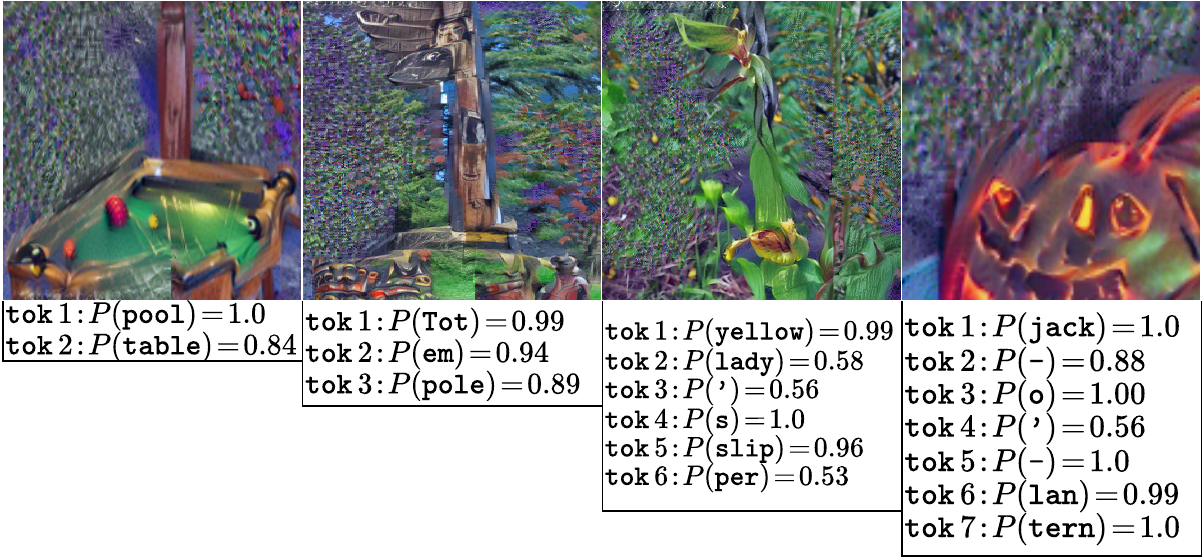}
    \captionof{figure}{\textbf{Examples over varying lengths} with Vicuna-13B.}
    \label{fig:target_length_ablations}
\end{minipage}$\;$
\begin{minipage}[b]{.25\textwidth}
    \centering
    \includegraphics[width=\linewidth]{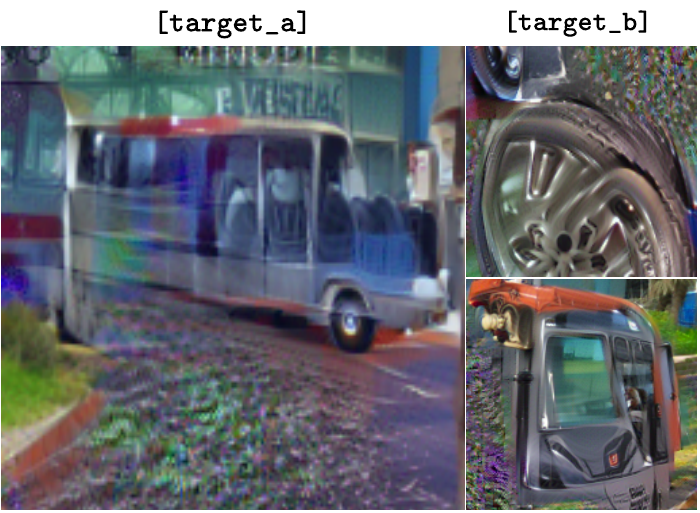}
    \captionof{figure}{\textbf{Combined ablations by changing from} \texttt{[target\_a]} \textbf{to} \texttt{[target\_b]} on Mistral-7B.}
    \label{fig:prompt_ablations}
\end{minipage}
\vspace{-2em}
\end{figure*}

\section{Results}
\label{sec:results}

\noindent
\textbf{Model Details}
We invert visual-instruct-tuned~\cite{li2024llava}  LLaMA3-8B~\cite{grattafiori2024llama}, Mistral-7B~\cite{jiang2023mistral}, and Vicuna-7/13B~\cite{zheng2023judging}. Models only run inference with the updatable input $\color{grdpurp}\widehat{\mathbf{v}}\color{black} \in \mathbb{R}^{3 \times 448 \times 448}$ initialized from a Gaussian $\color{grdpurp}\widehat{\mathbf{v}}\color{black}\sim\mathcal{N}(0, 1)$. We compute $\mathcal{L}_{base}$ statistics from~\cite{imagenet}. We use a 100-iteration warmup exponential lr scheduler starting from 0.05 over 5 gradient accumulation steps. A single run takes 15 min. on an L40s. Hyperparameters $\{a_1,a_2,a_3,\beta_1,\beta_2,\gamma_1,\gamma_2\}$ for all models are set with parallel hyperparameter tuning~\cite{sandha2020mango}. We run this once (3h) and use the same hyperparams throughout experiments. 

\noindent
\textbf{Evaluation Metrics}. We evaluate image and semantic quality using perceptual similarity~\cite{zhang2018lpips}, semantic alignment~\cite{hessel2022clipscore}, and image quality~\cite{heusel2018fid, salimans2016inception} metrics. Different \texttt{[target]} token lengths are quantitatively evaluated with BLEU~\cite{papineni2002bleu}, METEOR~\cite{banerjee2005meteor}, and ROUGE-L~\cite{chin2004rouge} across $|\color{grdpurp}\widehat{\mathbf{y}}\color{black}|$.

\noindent
\textbf{Main results}. We metrically assess synthesized images' quality in~\cref{tab:main1} reporting Fr\'echet inception distance (FID)~\cite{heusel2018fid} and Learned Perceptual Image Patch Similarity (LPIPS)~\cite{zhang2018lpips} between real~\cite{schuhmann2022laion} and synthesized images. The results show that the joint MIMIC objective is critical for obtaining understandable visualizations of features. MIMIC images are also better semantically aligned with respect to \texttt{[target]} tokens given the Inception Score (IS)~\cite{salimans2016inception} and CLIPScore (CScr)~\cite{hessel2022clipscore} performance.

\noindent
\textbf{Qualitative examples}. Alignment is also shown with qualitative examples in~\cref{fig:qualitative}. The images depict learned visual characteristics such as green-red feathers and black beak for \texttt{[macaw]}, bookshelves and study desks for \texttt{[library]}, and glass reflections for \texttt{[hourglass]}. The visualizations further allow exploration of learned correlations, such as the primary association of sport jerseys and shorts with \texttt{[rugby ball]} and prairies when optimizing \texttt{[barn]}.

\noindent
\textbf{Ablation studies}. We further ablate template $\mathbf{t}$ across vision tokens $\color{grdpurp} \widehat{\mathbf{v}}$ in~\cref{fig:context-ablations}. MIMIC can robustly visualize the main learned features, such as water reflections in \texttt{[dock]} and the dial plate in \texttt{[magnetic compass]}. MIMIC also effectively inverts multi-tokens outputs of varying \texttt{[target]} lengths $|\color{grdpurp}\widehat{\mathbf{y}}\color{black}|$, shown metrically in~\cref{fig:prompt_ablations} with qualitative examples in~\cref{fig:target_length_ablations}. Our approach further allows the discovery of learned attributes through Chain-of-Through prompting~\cite{wei2022chain}. In~\cref{fig:prompt_ablations} (left side) we first use template $\textbf{t}_a$: \texttt{What is shown in the image?} with \texttt{[minibus]} as target. Then (right side), we update to $\textbf{t}_b$: \texttt{A minivan is shown in the image. What's the main characteristic?} and infer \texttt{[wheels]} and \texttt{[windshield]} from Mistral-7B. In turn, we use these responses as \texttt{[targets]}. 
\vspace*{-0.6em}

\section{Conclusion}
\noindent
We propose MIMIC, a framework for visually inverting VLMs that combines inversion and feature alignment objectives with spatial alignment, image smoothness, and semantic realism regularizers. MIMIC can be applied across diverse models and settings to identify the main features learned by VLMs. We believe this first step for interpreting VLM representations is a promising direction towards understanding multi-modal encodings.

{
    \small
    \bibliographystyle{ieeenat_fullname}
    \bibliography{refs_s}

\begin{thebibliography}{44}
\providecommand{\natexlab}[1]{#1}
\providecommand{\url}[1]{\texttt{#1}}
\expandafter\ifx\csname urlstyle\endcsname\relax
  \providecommand{\doi}[1]{doi: #1}\else
  \providecommand{\doi}{doi: \begingroup \urlstyle{rm}\Url}\fi

\bibitem[Bach et~al.(2015)Bach, Binder, et~al.]{lrp}
Sebastian Bach, Alexander Binder, et~al.
\newblock On pixel-wise explanations for non-linear classifier decisions by layer-wise relevance propagation.
\newblock \emph{PLOS ONE}, 2015.

\bibitem[Banerjee and Lavie(2005)]{banerjee2005meteor}
Satanjeev Banerjee and Alon Lavie.
\newblock Meteor: An automatic metric for mt evaluation with improved correlation with human judgments.
\newblock In \emph{ACLw}, 2005.

\bibitem[Chen et~al.(2024)Chen, Vondrick, and Mao]{chen2024selfie}
Haozhe Chen, Carl Vondrick, and Chengzhi Mao.
\newblock Selfie: Self-interpretation of large language model embeddings.
\newblock In \emph{ICML}, 2024.

\bibitem[Chin-Yew(2004)]{chin2004rouge}
Lin Chin-Yew.
\newblock Rouge: A package for automatic evaluation of summaries.
\newblock In \emph{TSBOw}, 2004.

\bibitem[Chuang et~al.(2024)Chuang, Wang, et~al.]{chuang2024faithlm}
Yu-Neng Chuang, Guanchu Wang, et~al.
\newblock Faithlm: Towards faithful explanations for large language models.
\newblock \emph{arxiv:2402.04678}, 2024.

\bibitem[Deng et~al.(2009)Deng, Dong, et~al.]{imagenet}
Jia Deng, Wei Dong, et~al.
\newblock Imagenet: A large-scale hierarchical image database.
\newblock In \emph{CVPR}, 2009.

\bibitem[Dorszewski et~al.(2025)Dorszewski, Tětková, et~al.]{dorszewski2025colors}
Teresa Dorszewski, Lenka Tětková, et~al.
\newblock From colors to classes: Emergence of concepts in vision transformers.
\newblock \emph{arxiv:2503.24071}, 2025.

\bibitem[Dosovitskiy et~al.(2021)Dosovitskiy, Beyer, Kolesnikov, et~al.]{dosovitskiy2021imageworth16x16words}
Alexey Dosovitskiy, Lucas Beyer, Alexander Kolesnikov, et~al.
\newblock An image is worth 16x16 words: Transformers for image recognition at scale.
\newblock \emph{arxiv:2010.11929}, 2021.

\bibitem[Esser et~al.(2024)Esser, Kulal, Blattmann, Entezari, M{\"u}ller, Saini, Levi, Lorenz, Sauer, Boesel, et~al.]{esser2024scaling}
Patrick Esser, Sumith Kulal, Andreas Blattmann, Rahim Entezari, Jonas M{\"u}ller, Harry Saini, Yam Levi, Dominik Lorenz, Axel Sauer, Frederic Boesel, et~al.
\newblock Scaling rectified flow transformers for high-resolution image synthesis.
\newblock In \emph{ICML}, 2024.

\bibitem[Fel et~al.(2023)Fel, Picard, et~al.]{fel2023craft}
Thomas Fel, Agustin Picard, et~al.
\newblock Craft: Concept recursive activation factorization for explainability.
\newblock In \emph{CVPR}, 2023.

\bibitem[Gandelsman et~al.(2025)Gandelsman, Efros, and Steinhardt]{gandelsman2025secondorder}
Yossi Gandelsman, Alexei~A. Efros, and Jacob Steinhardt.
\newblock Interpreting the second-order effects of neurons in clip.
\newblock In \emph{ICLR}, 2025.

\bibitem[Ghandeharioun et~al.(2024)Ghandeharioun, Caciularu, et~al.]{ghandeharioun2024patchscopes}
Asma Ghandeharioun, Avi Caciularu, et~al.
\newblock Patchscopes: A unifying framework for inspecting hidden representations of language models.
\newblock In \emph{ICML}, 2024.

\bibitem[Ghiasi et~al.(2022)Ghiasi, Kazemi, et~al.]{ghiasi2022plugin}
Amin Ghiasi, Hamid Kazemi, et~al.
\newblock Plug-in inversion: Model-agnostic inversion for vision with data augmentations.
\newblock In \emph{ICML}, 2022.

\bibitem[Golovanevsky et~al.(2025)Golovanevsky, Rudman, et~al.]{golovanevsky2025vlmsnotice}
Michal Golovanevsky, William Rudman, et~al.
\newblock What do vlms notice? a mechanistic interpretability pipeline for gaussian-noise-free text-image corruption and evaluation.
\newblock In \emph{ACL}, 2025.

\bibitem[Grattafiori et~al.(2024)Grattafiori, Dubey, Jauhri, Pandey, Kadian, Al-Dahle, Letman, Mathur, Schelten, Vaughan, et~al.]{grattafiori2024llama}
Aaron Grattafiori, Abhimanyu Dubey, Abhinav Jauhri, Abhinav Pandey, Abhishek Kadian, Ahmad Al-Dahle, Aiesha Letman, Akhil Mathur, Alan Schelten, Alex Vaughan, et~al.
\newblock The llama 3 herd of models.
\newblock \emph{arXiv:2407.21783}, 2024.

\bibitem[Grosse et~al.(2023)Grosse, Bae, et~al.]{grosse2023influencefunctions}
Roger Grosse, Juhan Bae, et~al.
\newblock Studying large language model generalization with influence functions.
\newblock \emph{arxiv:2308.03296}, 2023.

\bibitem[Hatamizadeh et~al.(2022)Hatamizadeh, Yin, et~al.]{hatamizadeh2022gradvit}
Ali Hatamizadeh, Hongxu Yin, et~al.
\newblock Gradvit: Gradient inversion of vision transformers.
\newblock In \emph{CVPR}, 2022.

\bibitem[Hessel et~al.(2022)Hessel, Holtzman, et~al.]{hessel2022clipscore}
Jack Hessel, Ari Holtzman, et~al.
\newblock Clipscore: A reference-free evaluation metric for image captioning.
\newblock In \emph{EMNLPS}, 2022.

\bibitem[Heusel et~al.(2017)Heusel, Ramsauer, Unterthiner, Nessler, and Hochreiter]{heusel2018fid}
Martin Heusel, Hubert Ramsauer, Thomas Unterthiner, Bernhard Nessler, and Sepp Hochreiter.
\newblock Gans trained by a two time-scale update rule converge to a local nash equilibrium.
\newblock \emph{NeurIPS}, 2017.

\bibitem[Jiang et~al.(2023)Jiang, Sablayrolles, Mensch, Bamford, Chaplot, Casas, Bressand, Lengyel, Lample, Saulnier, et~al.]{jiang2023mistral}
Albert~Q Jiang, A Sablayrolles, A Mensch, C Bamford, D~Singh Chaplot, Ddl Casas, F Bressand, G Lengyel, G Lample, L Saulnier, et~al.
\newblock Mistral 7b. arxiv.
\newblock \emph{arXiv:2310.06825}, 2023.

\bibitem[Li et~al.(2024)Li, Zhang, Zhang, Zhang, Li, Li, Ma, and Li]{li2024llava}
Feng Li, Renrui Zhang, Hao Zhang, Yuanhan Zhang, Bo Li, Wei Li, Zejun Ma, and Chunyuan Li.
\newblock Llava-next-interleave: Tackling multi-image, video, and 3d in large multimodal models.
\newblock \emph{arXiv:2407.07895}, 2024.

\bibitem[Meng et~al.(2023)Meng, Bau, et~al.]{meng2023factualassociations}
Kevin Meng, David Bau, et~al.
\newblock Locating and editing factual associations in gpt.
\newblock In \emph{NeurIPS}, 2023.

\bibitem[Nguyen et~al.(2017)Nguyen, Clune, et~al.]{nguyen2017plug}
Anh Nguyen, Jeff Clune, et~al.
\newblock Plug \& play generative networks: Conditional iterative generation of images in latent space.
\newblock In \emph{CVPR}, 2017.

\bibitem[Oikarinen and Weng(2024)]{oikarinen2024linear}
Tuomas Oikarinen and Tsui-Wei Weng.
\newblock Linear explanations for individual neurons.
\newblock In \emph{ICML}, 2024.

\bibitem[Papineni et~al.(2002)Papineni, Roukos, et~al.]{papineni2002bleu}
Kishore Papineni, Salim Roukos, et~al.
\newblock Bleu: a method for automatic evaluation of machine translation.
\newblock In \emph{ACL}, 2002.

\bibitem[Raghu et~al.(2021)Raghu, Unterthiner, et~al.]{raghu2021vision}
Maithra Raghu, Thomas Unterthiner, et~al.
\newblock Do vision transformers see like convolutional neural networks?
\newblock In \emph{NeurIPS}, 2021.

\bibitem[Salimans et~al.(2016)Salimans, Goodfellow, et~al.]{salimans2016inception}
Tim Salimans, Ian Goodfellow, et~al.
\newblock Improved techniques for training gans.
\newblock \emph{NeurIPS}, 2016.

\bibitem[Sandha et~al.(2020)Sandha, Aggarwal, Fedorov, and Srivastava]{sandha2020mango}
Sandeep~Singh Sandha, Mohit Aggarwal, Igor Fedorov, and Mani Srivastava.
\newblock Mango: A python library for parallel hyperparameter tuning.
\newblock In \emph{ICASSP}, 2020.

\bibitem[Schuhmann et~al.(2022)Schuhmann, Beaumont, Vencu, Gordon, Wightman, Cherti, Coombes, Katta, Mullis, Wortsman, et~al.]{schuhmann2022laion}
Christoph Schuhmann, Romain Beaumont, Richard Vencu, Cade Gordon, Ross Wightman, Mehdi Cherti, Theo Coombes, Aarush Katta, Clayton Mullis, Mitchell Wortsman, et~al.
\newblock Laion-5b: An open large-scale dataset for training next generation image-text models.
\newblock \emph{NeurIPS}, 2022.

\bibitem[Selvaraju et~al.(2017)Selvaraju, Cogswell, et~al.]{gradcam}
Ramprasaath~R Selvaraju, Michael Cogswell, et~al.
\newblock Grad-cam: Visual explanations from deep networks via gradient-based localization.
\newblock In \emph{ICCV}, 2017.

\bibitem[Shrikumar et~al.(2017)Shrikumar, Greenside, and Kundaje]{shrikumar2019learning}
Avanti Shrikumar, Peyton Greenside, and Anshul Kundaje.
\newblock Learning important features through propagating activation differences.
\newblock In \emph{ICML}, 2017.

\bibitem[Stergiou(2021)]{stergiou2021mind}
Alexandros Stergiou.
\newblock The mind’s eye: Visualizing class-agnostic features of cnns.
\newblock In \emph{ICIP}, 2021.

\bibitem[Stergiou and Deligiannis(2023)]{stergiou2023leaping}
Alexandros Stergiou and Nikos Deligiannis.
\newblock Leaping into memories: Space-time deep feature synthesis.
\newblock In \emph{ICCV}, 2023.

\bibitem[Sundararajan et~al.(2017)Sundararajan, Taly, and Yan]{sundararajan2017axiomatic}
Mukund Sundararajan, Ankur Taly, and Qiqi Yan.
\newblock Axiomatic attribution for deep networks.
\newblock In \emph{ICML}, 2017.

\bibitem[Tao et~al.(2024)Tao, Huang, et~al.]{tao2024probing}
Mingxu Tao, Quzhe Huang, et~al.
\newblock Probing multimodal large language models for global and local semantic representations.
\newblock In \emph{LREC-COLING}, 2024.

\bibitem[Thasarathan et~al.(2025)Thasarathan, Forsyth, et~al.]{thasarathan2025universal}
Harrish Thasarathan, Julian Forsyth, et~al.
\newblock Universal sparse autoencoders: Interpretable cross-model concept alignment.
\newblock In \emph{ICML}, 2025.

\bibitem[Wang et~al.(2020)Wang, Wang, et~al.]{wang2020scorecam}
Haofan Wang, Zifan Wang, et~al.
\newblock Score-cam: Score-weighted visual explanations for convolutional neural networks.
\newblock In \emph{CVPRw}, 2020.

\bibitem[Wei et~al.(2022)Wei, Wang, Schuurmans, Bosma, Xia, Chi, Le, Zhou, et~al.]{wei2022chain}
Jason Wei, Xuezhi Wang, Dale Schuurmans, Maarten Bosma, Fei Xia, Ed Chi, Quoc~V Le, Denny Zhou, et~al.
\newblock Chain-of-thought prompting elicits reasoning in large language models.
\newblock \emph{NeurIPS}, 2022.

\bibitem[Yin et~al.(2020)Yin, Molchanov, et~al.]{yin2020deepinversion}
Hongxu Yin, Pavlo Molchanov, et~al.
\newblock Dreaming to distill: Data-free knowledge transfer via deepinversion.
\newblock In \emph{CVPR}, 2020.

\bibitem[Yosinski et~al.(2015)Yosinski, Clune, et~al.]{yosinski2015understanding}
Jason Yosinski, Jeff Clune, et~al.
\newblock Understanding neural networks through deep visualization.
\newblock In \emph{ICMLw}, 2015.

\bibitem[Zhang et~al.(2018)Zhang, Isola, et~al.]{zhang2018lpips}
Richard Zhang, Phillip Isola, et~al.
\newblock The unreasonable effectiveness of deep features as a perceptual metric.
\newblock In \emph{CVPR}, 2018.

\bibitem[Zhao et~al.(2024)Zhao, Wang, et~al.]{zhao2025gradeclip}
Chenyang Zhao, Kun Wang, et~al.
\newblock Grad-eclip: Gradient-based visual and textual explanations for clip.
\newblock In \emph{ICML}, 2024.

\bibitem[Zheng et~al.(2023)Zheng, Chiang, Sheng, Zhuang, Wu, Zhuang, Lin, Li, Li, Xing, et~al.]{zheng2023judging}
Lianmin Zheng, Wei-Lin Chiang, Ying Sheng, Siyuan Zhuang, Zhanghao Wu, Yonghao Zhuang, Zi Lin, Zhuohan Li, Dacheng Li, Eric Xing, et~al.
\newblock Judging llm-as-a-judge with mt-bench and chatbot arena.
\newblock \emph{NeurIPS}, 2023.

\bibitem[Zhou et~al.(2016)Zhou, Khosla, et~al.]{zhou2015learning}
Bolei Zhou, Aditya Khosla, et~al.
\newblock Learning deep features for discriminative localization.
\newblock In \emph{CVPR}, 2016.

\end{thebibliography}
}


\clearpage
\setcounter{page}{1}
\setcounter{equation}{0}
\setcounter{table}{0}
\setcounter{figure}{0}
\AppendixA
\twocolumn[{%
\renewcommand\twocolumn[1][]{#1}%
\maketitlesupplementary
\begin{center}
    \vspace{-1em} 
    \centering
    \includegraphics[width=\textwidth]{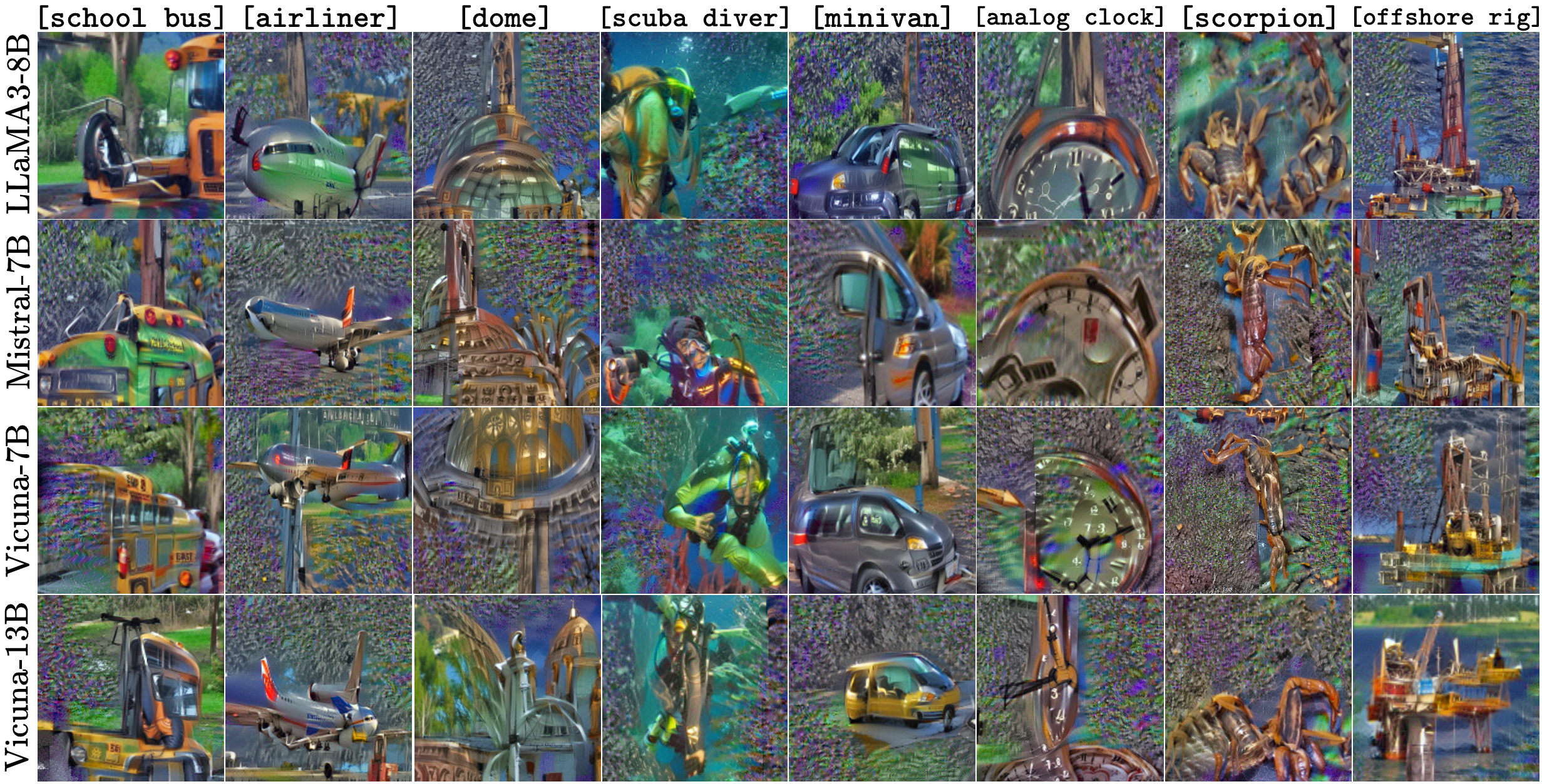}
    \captionof{figure}{\textbf{Qualitative examples of synthesized features} across \texttt{[target]} tokens.}
    \label{fig:qualitative_extra}
\end{center}%
}]
\section{Additional qualitative results}
\label{appendix:additional_qualitative}

We demonstrated qualitative results of inverted \texttt{[target]} VLM tokens in~\cref{sec:results}. Supplementary to~\cref{fig:qualitative}, we further visualize additional examples across VLMs in~\cref{fig:qualitative_extra}. As shown, MIMIC synthesizes coherent features across models with various target tokens. Descriptive VLM features learned for target semantics are often based on distinct shapes such as the examples for \texttt{[airliner]} and \texttt{[offshore rig]}. Positive correlations between materials and colors are also learned for instances such as \texttt{[school bus]}, \texttt{[dome]}, and \texttt{[minivan]}. VLMs also learn environment associations, such as oxygen bubbles, for \texttt{[scuba diving]}.

\begin{figure}[!ht]
    \centering
    \includegraphics[width=\linewidth]{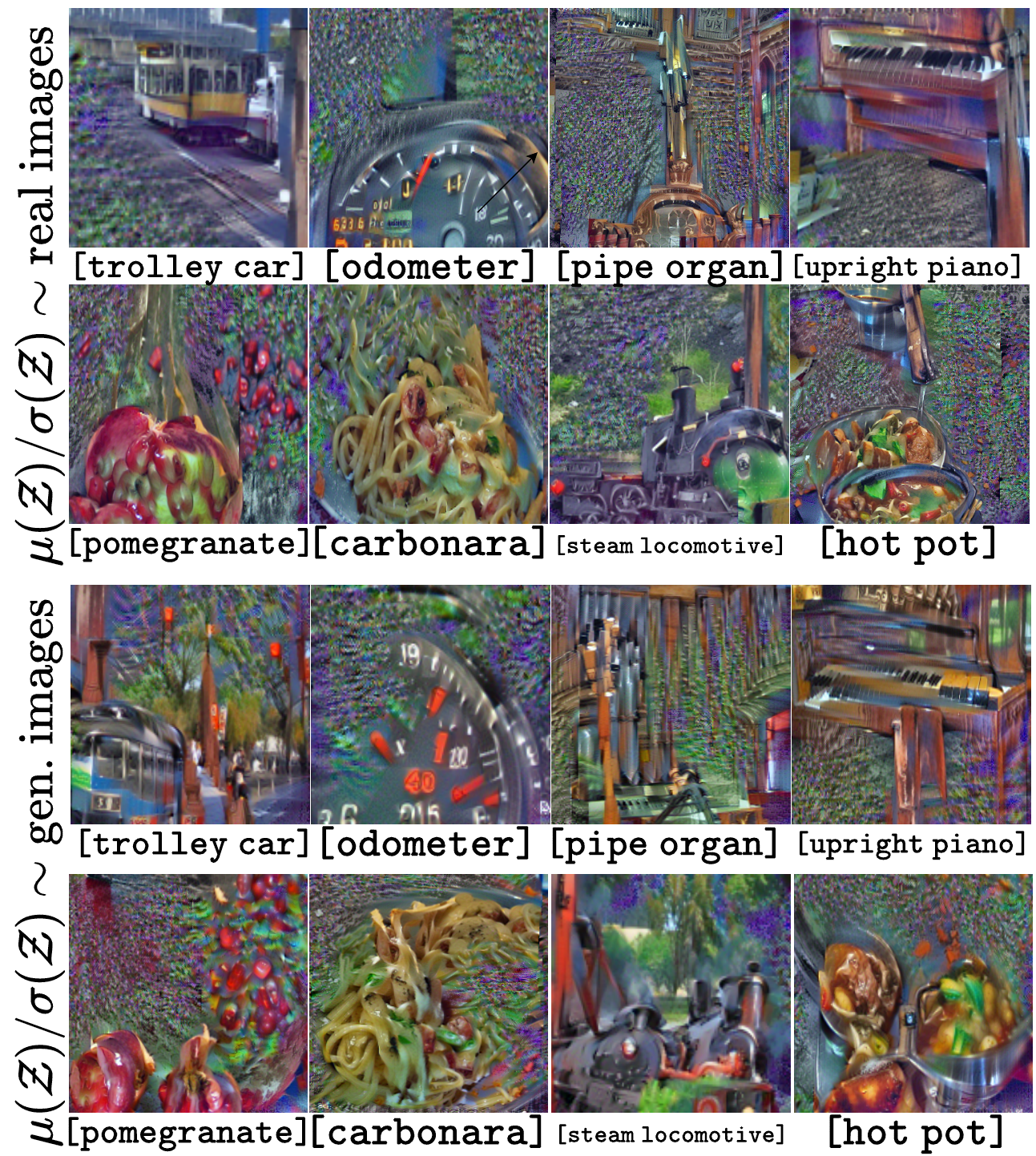}
    \vspace*{-1.2em}
    \caption{\textbf{Comparisons between statistics from real and generated images}. (top) includes base feature loss as in~\cref{sec:method}. (bottom) computes base loss from SD3-M~\cite{esser2024scaling} image embeddings.}
    \label{fig:real2dif_extra}
    \vspace{-1em}
\end{figure}

\section{Generated base feature priors}
\label{appendix:base_feat}

As described in~\cref{sec:method}, internal vision encoder layer statistics $\mu(\mathcal{Z}_l)$,$\sigma(\mathcal{Z}_l)$ can be approximated without sourced images. For this, we instead approximate \texttt{[target]} manifold's distribution by sourcing images generated with SD3-M~\cite{esser2024scaling} with prompt template \texttt{A photorealistic image of [target]} as the condition.~\cref{fig:real2dif_extra} demonstrates that statistics from generated images do not result in a significant quality drop, with images still including visually distinct features.

\end{document}